\documentclass{article}
\usepackage{ijcai24}

\usepackage{times}
\usepackage{soul}
\usepackage{url}
\usepackage[hidelinks]{hyperref}
\usepackage[utf8]{inputenc}
\usepackage[small]{caption}
\usepackage{graphicx}
\usepackage{amsmath}
\usepackage{amsthm}
\usepackage{booktabs}
\usepackage{algorithm}
\usepackage{algorithmic}
\usepackage[switch]{lineno}
\graphicspath{{figs}}
\usepackage{pifont}
\usepackage{tabularx}
\usepackage{natbib}
\usepackage{enumitem}
\usepackage[dvipsnames]{xcolor}
\definecolor{my_green}{rgb}{0.429, 0.850, 0.208}
\definecolor{my_blue}{rgb}{0.196, 0.510, 0.884}
\title{VIALM: A Survey and Benchmark of \\
Visually Impaired Assistance with Large Models}

\author{
Yi Zhao$^1$\and
Yilin Zhang$^1$\and
Rong Xiang$^{1}$\and
Jing Li$^{1}$\And
Hillming Li$^{2}$\\
\affiliations
$^1$Department of Computing, The Hong Kong Polytechnic University\and
$^2$Hillming Smartech Ltd.
\emails
yi-yi-yi.zhao@connect.polyu.hk\and
li@hillming.com \\
\{yilin-jason.zhang, rong-chris.xiang, jing-amelia.li\}@polyu.edu.hk
}

\begin{document}

\maketitle
\begin{abstract}
	Visually Impaired Assistance (VIA) aims to automatically help the visually impaired (VI) handle daily activities.
	The advancement of VIA primarily depends on developments in Computer Vision (CV) and Natural Language Processing (NLP), both of which exhibit cutting-edge paradigms with large models (LMs).
	Furthermore, LMs have shown exceptional multimodal abilities to tackle challenging physically-grounded tasks such as embodied robots. 
	To investigate the potential and limitations of state-of-the-art (SOTA) LMs' capabilities in VIA applications, we present an extensive study for the task of VIA with LMs (\textbf{VIALM}). 
	In this task, given an \textit{image} illustrating the physical environments and a \textit{linguistic request} from a VI user, VIALM aims to output step-by-step \textit{guidance} to assist the VI user in fulfilling the request grounded in the environment. 
	The study consists of a survey reviewing recent LM research and benchmark experiments examining selected LMs' capabilities in VIA.
	The results indicate that while LMs can potentially benefit VIA, their output cannot be well \textit{environment-grounded} (i.e., 25.7\% GPT-4's responses) and lacks \textit{fine-grained} guidance (i.e., 32.1\% GPT-4's responses).
\end{abstract}
\section{Introduction}
%VIA meaning and past studies
The term \textit{Visually Impaired} (VI) refers to individuals with partial or complete blindness.  
It includes approximately 2.2 billion people worldwide.\footnote{\url{https://www.who.int/publications/i/item/9789241516570}}
To improve their lives, numerous previous studies have focused on \textit{Visually Impaired Assistance} (VIA).
It aims to provide support, tools, and services to aid the VI in their daily activities and  promote their independence.
The mainstream VIA research takes advantage of the advanced CV and NLP technology.
For example, BrowseWithMe \citep{DBLP:conf/assets/StanglKJYGG18} can talk with VI users via NLP to assist online shopping;
ReCog \citep{DBLP:conf/chi/Ahmetovic0OIKA20} can help VI users understand the surrounding objects via CV.

%Drawback od Small models, LMs' achievements
To date, the predominant VIA efforts have been made with small models.
However, large models (LMs) have recently transformed the landscape of CV, NLP, and multimodal research \citep{DBLP:conf/nips/BrownMRSKDNSSAA20}.
For instance, in NLP, a Large Language Model (LLM) ChatGPT \citep{DBLP:conf/nips/Ouyang0JAWMZASR22} showed superior cross-task language skills.
Moreover, a large Visual-Language Model (VLM) GPT-4 \citep{DBLP:journals/corr/abs-2303-08774} exhibited  human-equivalent abilities in non-trivial multimodal tasks.
More recently, a LLM-based embodied agent named PaLM-E \citep{DBLP:conf/icml/DriessXSLCIWTVY23} further grounded the vision and language capabilities to interact with physical environments.

%VIALM task definition
\begin{figure}
	\centering
	\includegraphics[angle=0,width=0.5\textwidth]{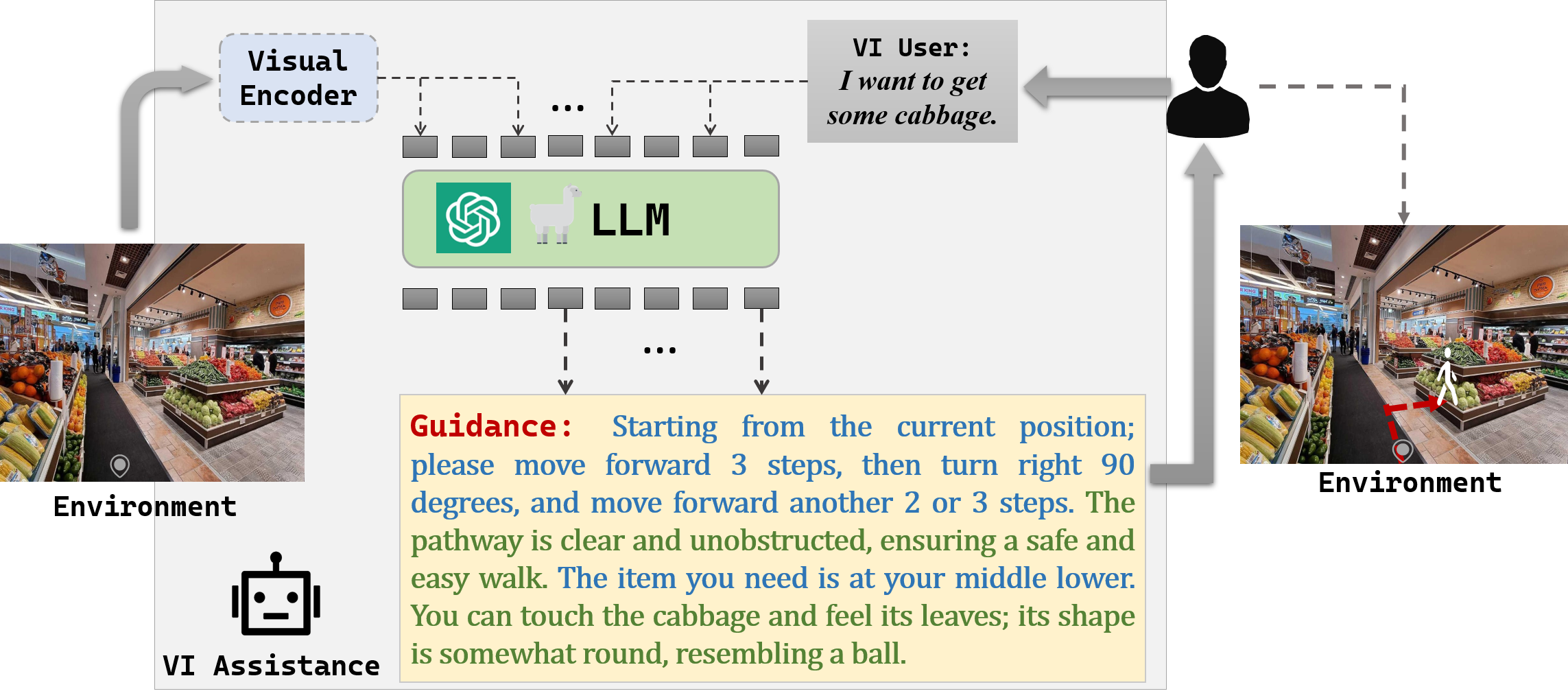}
	\caption{A sample input and output of VIALM. Its input is a pair of a visual image of the environment (the left image) and a user request in language (the \colorbox{lightgray}{grey} box). The yellow box shows the output guidance for VI users to complete the request within the environment (the right image). The output should convey \textit{environment-grounded} information (\textcolor{my_blue}{blue} words) and \textit{fine-grained} (\textcolor{my_green}{green} words) guidance, integrating \textit{tactile} support (the last sentence) for VI users.
	}
	\label{VIALM Task Illustration}
\end{figure}
Viewing this revolution, we raise a crucial question: \textit{can LMs transform the landscape of VIA}?
We consequently propose a task of \textbf{Visual Impaired Assistance with Language Models (VIALM)}.
The input of VIALM is an image illustrating the physical environment and a language request from a VI user about a task to complete in the environment.
The output is a step-by-step \emph{environment-grounded} guidance to assist the VI user in fulfilling their request.
This guidance need to be easy to follow (\textit{fine-grained}) for these users.
Particularly, it is crucial to involve \emph{tactile} guidance as VI users heavily rely on tactile cognition to complement vision loss \citep{ottink2022cognitive}.
Figure \ref{VIALM Task Illustration} shows a sample input and output of VIALM.

%This work
To the best of our knowledge, \textit{our study is the first to investigate LMs in VIA extensively}.  
It consists of a thorough survey of important LMs' work, including LLMs, VLMs, and embodied agents, all of which hold potential benefits for VIA.
Subsequently, we present a fundamental benchmark for empirical studies by formulating the task following the Visual Question Answering (VQA) setup. 
It comprises 200 visual environment \textit{images} with paired \textit{questions} (user requests) and \textit{answers} (VIA guidance).
The images cover two types of environments: supermarket and home.
Based on the benchmark, we examined 6 SOTA VLMs, GPT-4, CogVLM \citep{DBLP:journals/corr/abs-2311-03079}, Qwen-VL \citep{DBLP:journals/corr/abs-2308-12966}, LLaVA \citep{DBLP:journals/corr/abs-2304-08485}, MiniGPT-v2 \citep{DBLP:journals/corr/abs-2310-09478}, and BLIVE \citep{DBLP:journals/corr/abs-2308-09936}, to study their zero-shot capabilities in VIA.

%Add Findings
Benchmark experiments identify two main limitations in current SOTA LMs: (1) inferiority to generate environment-grounded guidance (25.7\% of GPT-4's responses), and (2) a lack of fine-grained guidance (32.1\% of GPT-4's responses),
with a shortfall in integrating tactile sensation.
However, it is also revealed that visually-focused LMs excel in environment understanding, while those with advanced language models are superior in generating easy-to-follow guidance.
To overcome these limitations, we propose potential solutions: (1) improve visual capabilities to enhance environmental grounding, and (2) incorporate tactile modalities into language generation to produce more fine-grained guidance.
A promising direction for future research involves synergistically enhancing multimodal capabilities to boost overall effectiveness.

% contributions
For future developments, we have open-sourced the resources.\footnote{\url{https://github.com/YiyiyiZhao/VIALM}} The contributions of this work are threefold:

$\bullet$ We introduce a novel task, VIALM, to extensively investigate how LMs transform the VIA landscape.

$\bullet$ We thoroughly survey the important LM work applicable to VIA and construct the first VIALM benchmark.

$\bullet$ We experiment with our benchmark and draw practical findings regarding SOTA LMs' zero-shot VIA capabilities.

\section{Large Models}
\begin{figure}
	\centering
	\includegraphics[angle=0,width=0.5\textwidth]{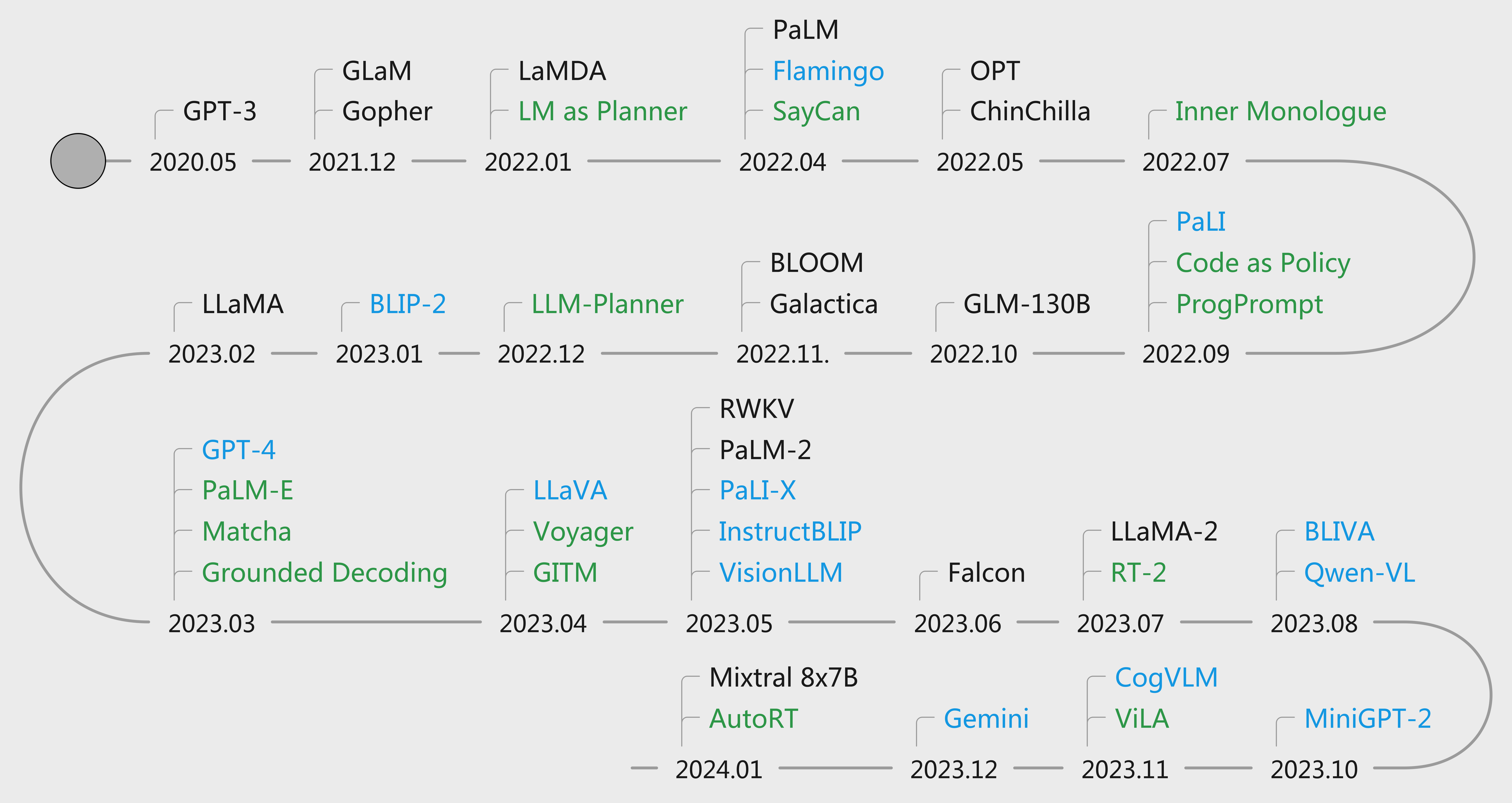}
	\caption{Timeline of LMs.
		Release times are based on the publication dates of their respective descriptive papers on arXiv.
		LLMs are marked in \textcolor{black}{black}, large VLMs in \textcolor{my_blue}{blue}, and embodied agents in \textcolor{my_green}{green}. It is observed the years 2021 and 2022 saw LLM advancements, followed by a 2023 surge in VLMs and embodied agent development.}
	\label{LM Timeline}
\end{figure}

Over the past two years, LMs have dramatically reshaped the fields of CV, NLP, and multimodal research.
This section offers a comprehensive and current survey of LMs, focusing on 43 notable examples, chronologically depicted in Figure \ref{LM Timeline}.
It is observed that the years 2021 and 2022 marked advancements in LLMs, followed by an upsurge in the development of VLMs and embodied agents in 2023 and 2024.
\subsection{Large Language Models}
\begin{table*}
	\centering
	\scriptsize
	\begin{tabularx}{\textwidth}{lrrrrl}
		\toprule
		\textbf{Name} & \textbf{Release Time} & \textbf{Pretraining Task} & \textbf{Model Architecture} & \textbf{\# Maximum Parameters} & \textbf{Instruction-tuned Variants} \\
		\midrule
		GPT-3 \citep{DBLP:conf/nips/BrownMRSKDNSSAA20} & 2020.05 & \ding{192} & \ding{192} & 175B & \begin{tabular}[c]{@{}l@{}}InstructGPT, \\ GPT-3.5\end{tabular} \\ \hline
		GLaM \citep{DBLP:conf/icml/DuHDTLXKZYFZFBZ22}& 2021.12 & \ding{192} & \ding{193} & 1.2T & / \\
		Gopher \citep{DBLP:journals/corr/abs-2112-11446} & 2021.12 & \ding{192} & \ding{192} & 280B & / \\
		LaMDA \citep{DBLP:journals/corr/abs-2201-08239}& 2022.01 & \ding{192} & \ding{192} & 137B & Bard \citep{manyika2023overview} \\
		PaLM \citep{DBLP:journals/jmlr/ChowdheryNDBMRBCSGSSTMRBTSPRDHPBAI23} & 2022.04 & \ding{192} & \ding{192} & 540B & Flan-PaLM \citep{DBLP:journals/corr/abs-2210-11416}\\
		Chinchilla \citep{DBLP:journals/corr/abs-2203-15556} & 2022.05 & \ding{192} & \ding{192} & 70B & / \\
		OPT \citep{DBLP:journals/corr/abs-2205-01068}& 2022.05 & \ding{192} & \ding{192} & 175B & OPT-IML \citep{DBLP:journals/corr/abs-2212-12017}\\
		GLM-130B \citep{DBLP:conf/iclr/ZengLDWL0YXZXTM23}& 2022.10 & \ding{193} & \ding{194} & 130B & ChatGLM \\
		BLOOM \citep{DBLP:journals/corr/abs-2211-05100} & 2022.11 & \ding{192} & \ding{192} & 176B & BLOOMZ \citep{DBLP:conf/acl/MuennighoffWSRB23}\\
		Galactica \citep{DBLP:journals/corr/abs-2211-09085}& 2022.11 & \ding{192} & \ding{192} & 120B & Evol-Instruct\\ \hline
		LLaMA \citep{DBLP:journals/corr/abs-2302-13971} & 2023.02 & \ding{192} & \ding{192} & 65B & \begin{tabular}[c]{@{}l@{}}Alpaca \citep{taori2023alpaca},\\ WizardLM,\\ Vicuna \citep{chiang2023vicuna}\end{tabular} \\ \hline
		RWKV \citep{DBLP:conf/emnlp/PengAAAABCCCDDG23} & 2023.05 & \ding{192} & \ding{195} & 14B & RWKV-4 Raven \\
		PaLM-2 \citep{DBLP:journals/corr/abs-2305-10403}& 2023.05 & \ding{194} & \ding{196} & / & / \\ \hline
		LLaMA-2 \citep{DBLP:journals/corr/abs-2307-09288}& 2023.07 & \ding{192} & \ding{192} & 70B & \begin{tabular}[c]{@{}l@{}}LLaMA2-Chat,\\ OpenChat V2\end{tabular} \\ \hline
		Falcon \citep{DBLP:journals/corr/abs-2311-16867}& 2023.11 & \ding{192} & \ding{192} & 180B & Falcon-instruct \\
		Mixtral 8x7B  \citep{DBLP:journals/corr/abs-2401-04088}& 2024.01 & \ding{192} & \ding{193} & 13B & Mixtral 8x7B–Instruct \\
		\bottomrule
	\end{tabularx}
	\caption{Summary of pretrained LLMs and their instruction-tuned variants, arranged chronologically from the earliest to the most recent releases.
		Categorized by pretraining tasks, symbols \ding{192}, \ding{193}, and \ding{194} denote language modeling \protect\citep{radford2019language}, autoregressive blank infilling, and mixed objectives, respectively.
		Categorized by model architectures, symbols \ding{192}, \ding{193}, \ding{194} and \ding{195} represent transformer decoder \protect\citep{radford2018improving}, MOE decoder, bidirectional GLM, RWKV architecture, and encoder-decoder transformer, respectively.
		The symbol "/" is used when specific information is not explicitly available.
		All these models have a maximum parameter count exceeding 10 billion.
		The majority of these LLMs were released in the past two years.
		It is evident that many instruction-tuned models are developed using LLaMA.}
	\label{tab:LLM}
\end{table*}
%Why survey LLMs
Language models are neural networks with text input and output, initially designed for NLP tasks.
Large Language Models (LLMs) stand out by their extensive scale, as measured by network parameters.
A key feature of LLMs is their emergent abilities \citep{DBLP:journals/tmlr/WeiTBRZBYBZMCHVLDF22}, which have been leveraged in reasoning and decision-making \citep{DBLP:conf/iclr/YaoZYDSN023}.
We focus on models with more than 10 billion parameters, in line with the definition in  \citet{DBLP:journals/corr/abs-2303-18223}.
This survey categorizes LLMs based on architectures  and pretraining tasks and investigates instruction-tuned LLMs, summarized in Table \ref{tab:LLM}.

\paragraph{Pretraining Tasks.} The predominant pretraining task for LLMs is language modeling \citep{radford2019language}, which predicts the next tokens in an autoregressive manner.
The success of GPT series \citep{DBLP:conf/nips/BrownMRSKDNSSAA20} has demonstrated this approach's effectiveness, especially when the language model scale is significantly increased and the training corpus is extensively expanded.
Consequently, a substantial number of LLMs have adopted it, including GLaM \citep{DBLP:conf/icml/DuHDTLXKZYFZFBZ22}, Gopher \citep{DBLP:journals/corr/abs-2112-11446}, LaMDA \citep{DBLP:journals/corr/abs-2201-08239}, PaLM \citep{DBLP:journals/jmlr/ChowdheryNDBMRBCSGSSTMRBTSPRDHPBAI23}, Chinchilla \citep{DBLP:journals/corr/abs-2203-15556}, OPT \citep{DBLP:journals/corr/abs-2205-01068}, BLOOM \citep{DBLP:journals/corr/abs-2211-05100}, Galactica \citep{DBLP:journals/corr/abs-2211-09085}, LLaMA \citep{DBLP:journals/corr/abs-2302-13971}, RWKV \citep{DBLP:conf/emnlp/PengAAAABCCCDDG23}, LLaMA-2 \citep{DBLP:journals/corr/abs-2307-09288}, Falcon \citep{DBLP:journals/corr/abs-2311-16867}, and Mixtral \citep{DBLP:journals/corr/abs-2401-04088}. 
Meanwhile, some LLMs have implemented mixed schemes to enhance language understanding.
For example, GLM-130B has introduced autoregressive blank infilling \citep{DBLP:conf/iclr/ZengLDWL0YXZXTM23}, and more recently, PaLM-2 has adopted a mixture of objectives \citep{DBLP:journals/corr/abs-2305-10403}.
\paragraph{Model Architectures.} The model architectures generally align with pretraining tasks.
Predominantly, LLMs utilize the transformer decoder architecture, which synergizes well with language modeling pretraining.
Some models have embraced this architecture, including GPT-3, Gopher, LaMDA, PaLM, Chinchilla, OPT, BLOOM, Galactica, LLaMA, LLaMA-2, and Falcon.
Additionally, there are variations in model architecture.
For instance, GLaM and Mixtral employ the Mixture of Experts (MOE) approach, GLM-130B features a bidirectional design, RWKV introduces their unique architecture, and PaLM-2 has adopted the encoder-decoder architecture.
\paragraph{Instruction-tuned LLMs.}Instruction tuning employs a limited instruction dataset combined with multitask fine-tuning, showing advantages in aligning LLMs with human intentions \citep{DBLP:conf/iclr/WeiBZGYLDDL22, li2023shot}.
Many fundamental LLMs have related instruction-tuned variants.
One category of variants, with ChatGPT being the most renowned, is those capable of functioning in a conversational manner.
Additionally, LLaMA and LLaMA-2 have become prominent fundamental models for creating instruction-tuned LLMs.
Notable tuned examples include Alpaca \citep{taori2023alpaca}, Vicuna \citep{chiang2023vicuna}, LLaMA2-Chat, among others.
\subsection{Large Vision-Language Models}
Large Vision-Language Models (VLMs) process both visual and language inputs, enabling them to handle multimodal tasks such as image captioning and VQA.
In the context of VIA, we focus on VLMs that generate linguistic outputs, ensuring accessibility for VI users.
A typical VLM architecture consists of three key components: a visual encoder, a LLM, and a vision-language connector \citep{DBLP:journals/corr/abs-2304-08485}. This survey categorizes VLMs according to their model components and training/tuning strategies, as summarized in Table \ref{tab:VLM}.
\begin{table*}
	\scriptsize
	\begin{tabularx}{\textwidth}{lrllll}
		\toprule
		\textbf{Name} & \textbf{Release Time} & \textbf{Visual Encoder} & \textbf{LLM} & \textbf{Connector} & \textbf{Training/Tuning} \\
		\midrule
		\begin{tabular}[c]{@{}l@{}}Flamingo\\ \citep{DBLP:conf/nips/AlayracDLMBHLMM22}\end{tabular} & 2022.04 & NFNet & Chinchilla & Cross-attention dense layers & Pretraining \\ \hline
		\begin{tabular}[c]{@{}l@{}}PaLI \\ \citep{DBLP:conf/iclr/Chen0CPPSGGMB0P23}\end{tabular} & 2022.09 & \begin{tabular}[c]{@{}l@{}}ViT-e\\ \citep{DBLP:conf/cvpr/Zhai0HB22}\end{tabular} & mT5 \citep{DBLP:journals/jmlr/RaffelSRLNMZLL20}  & / & Multi-task Pretraining \\ \hline
		\begin{tabular}[c]{@{}l@{}}BLIP-2 \\ \citep{DBLP:conf/icml/0008LSH23}\end{tabular} & 2023.01 & \begin{tabular}[c]{@{}l@{}}1. ViT-L/14 \\ \citep{DBLP:conf/icml/RadfordKHRGASAM21},  \\ 2. ViT-g/14\\  \citep{DBLP:conf/cvpr/FangWXSWW0WC23}\end{tabular} & \begin{tabular}[c]{@{}l@{}}1. OPT\\ 2. FlanT5\\ \end{tabular} & Q-former & Multi-task Pretraining \\ \hline
		\begin{tabular}[c]{@{}l@{}}GPT-4 \\ \citep{DBLP:journals/corr/abs-2303-08774}\end{tabular} & 2023.03 & / & GPT-4 & / & / \\ \hline
		\begin{tabular}[c]{@{}l@{}}LLaVa \\ \citep{DBLP:journals/corr/abs-2304-08485}\end{tabular} & 2023.04 & ViT-L/14 & LLaMA & Projection layer & \begin{tabular}[c]{@{}l@{}}1. Pretraining,\\ 2. Instruction tuning.\end{tabular} \\ \hline
		%\begin{tabular}[c]{@{}l@{}}MiniGPT-4\\ \citep{DBLP:journals/corr/abs-2304-10592}\end{tabular} & 2023.04 & \begin{tabular}[c]{@{}l@{}}Blip-2 ViT-g/14\\ and Q-Former\end{tabular} & Vicuna & Linear projection layer & \begin{tabular}[c]{@{}l@{}}1. Pretraining,\\ 2. Instruction tuning.\end{tabular} \\ \hline
		\begin{tabular}[c]{@{}l@{}}PaLI-X\\ \citep{DBLP:journals/corr/abs-2305-18565}\end{tabular} & 2023.05 & \begin{tabular}[c]{@{}l@{}}ViT-22B\\ \citep{DBLP:conf/icml/0001DMPHGSCGAJB23}\end{tabular} & UL2 & Projection layer & \begin{tabular}[c]{@{}l@{}}1. Multi-task pretraining,\\ 2. Task-specific fine-tuning.\end{tabular} \\ \hline
		\begin{tabular}[c]{@{}l@{}}InstructBLIP\\ \citep{DBLP:journals/corr/abs-2305-06500}\end{tabular} & 2023.05 & ViT-g/14 & \begin{tabular}[c]{@{}l@{}}1. FlanT5 \\ 2. Vicuna\end{tabular} & Q-former & \begin{tabular}[c]{@{}l@{}}1. Pretraining,\\ 2. Instruction tuning.\end{tabular} \\ \hline
		\begin{tabular}[c]{@{}l@{}}VisionLLM\\ \citep{DBLP:journals/corr/abs-2305-11175}\end{tabular} & 2023.05 & \begin{tabular}[c]{@{}l@{}}1. ResNet, \\ 2. InternImage-H\end{tabular} & Alpaca & \begin{tabular}[c]{@{}l@{}}BERT-Base and \\ Deformable DETR \citep{DBLP:conf/iclr/ZhuSLLWD21}\end{tabular} & Multi-task pretraining \\ \hline
		\begin{tabular}[c]{@{}l@{}}BLIVA\\ \citep{DBLP:journals/corr/abs-2308-09936}\end{tabular} & 2023.08 & ViT-g/14 & FlanT5 & Q-former and projection layer & \begin{tabular}[c]{@{}l@{}}1. Pretraining,\\ 2. Instruction-tuning.\end{tabular} \\ \hline
		\begin{tabular}[c]{@{}l@{}}Qwen-VL\\ \citep{DBLP:journals/corr/abs-2308-12966}\end{tabular} & 2023.08 & \begin{tabular}[c]{@{}l@{}}ViT\\ \citep{DBLP:conf/iclr/DosovitskiyB0WZ21}\end{tabular} & Qwen & Single-layer cross-attention module & \begin{tabular}[c]{@{}l@{}}1. Pretraining,\\ 2. Multi-task training,\\ 3. Conversation tuning.\end{tabular} \\ \hline
		\begin{tabular}[c]{@{}l@{}}MiniGPT-v2\\ \citep{DBLP:journals/corr/abs-2310-09478}\end{tabular} & 2023.10 & ViT-g/14 & LLaMA2-Chat & Linear projection layer & \begin{tabular}[c]{@{}l@{}}1. Pretraining,\\ 2. Multi-task training,\\ 3. Instruction tuning.\end{tabular} \\ \hline
		\begin{tabular}[c]{@{}l@{}}CogVLM\\ \citep{DBLP:journals/corr/abs-2311-03079}\end{tabular} & 2023.11 & \begin{tabular}[c]{@{}l@{}}Eva-clip ViT\\ \citep{DBLP:journals/corr/abs-2303-15389}\end{tabular} & Vicuna & MLP adapter & \begin{tabular}[c]{@{}l@{}}1. Pretraining;\\ 2. Multi-task training\end{tabular} \\ \hline
		\begin{tabular}[c]{@{}l@{}}Gemini \\ 
			\citep{DBLP:journals/corr/abs-2303-08774}\end{tabular} & 2023.12 & / & PaLM & / & / \\ 
		
		\bottomrule
	\end{tabularx}
	\caption{Summary of pretrained VLMs.
		A large VLM typically comprises a visual encoder, a LLM, and a connector.
		It is indicated that ViT is the most favored visual encoder, while LLMs primarily derive from the LLaMA family.
		The most common connectors are projection layers.
		Pretraining for VLMs generally encompasses three phases: foundational training, multi-task fine-tuning, and instructional tuning.
	}
	\label{tab:VLM}
\end{table*}

\paragraph{Visual Components.} Table \ref{tab:VLM} indicates that ViT is the most preferred visual encoder in current VLMs. Qwen-VL implements the standard ViT framework, whereas PaLI-X utilizes the highly scaled-up 22B variant.
The ViT-L/14 and ViT-g/14 models, both examined in the BLIP-2 study, are particularly popular.
LLaVA has adopted ViT-L/14, while InstructBLIP, BLIVA, and MiniGPT-v2 have selected ViT-g/14.
The ViT transformer architecture offers substantial benefits in visual encoding, including scalability in both model and data size \citep{DBLP:conf/icml/0001DMPHGSCGAJB23}, and a more effective integration of visual and language embeddings \citep{DBLP:conf/cvpr/FangWXSWW0WC23}.

\paragraph{LLM Components.}
During early VLM development, models such as BLIP-2 and InstructBLIP adopted the instruction-tuned FlanT5 as the LLM component.
Recently, the open-source, high-performing LLaMA family has drawn significant attention, with LLaMA-2's release likely to amplify this trend.
Many VLMs, including LLaVA, MiniGPT-v2, and CogVLM, have adopted language models from the LLaMA family, such as Vicuna, Alpaca, and LLaMA-2-Chat.

\paragraph{Cross-Modal Connectors.} BLIP-2 introduced the Q-Former component, which was subsequently adopted by InstructBLIP and BLIVA. 
However, the most common design for connectors currently is a linear projection layer or a multi-layer perceptron (MLP).
Models like LLaVA, PaLI-X, MiniGPT-v2, and CogVLM have adopted this design.
Similarly, Qwen-VL has adopted the linear layer connection approach with additional cross-attention mechanisms.
Comparing MLP with Q-Former, the LLaVA-v1.5 study \citep{DBLP:journals/corr/abs-2310-03744} suggests that the Q-Former may have shortcomings in effectively managing the output length of LLMs.

\paragraph{Training Methods.} VLM training typically goes through three phases.
Initially, the pretraining phase utilizes extensive image-text pairs from the Internet to equip the model with general multimodal capabilities.
Then, models undergo multi-task fine-tuning with more carefully curated datasets.
The final stage centers on instruction tuning with a limited amount of instructional or conversational data.
During these stages, the training components of the models may differ.
In the first stage, the connector may update its parameters.
However, models typically do not update the LLM parameters at this stage due to their substantial scale.
In contrast, many VLMs update the LLM parameters during the final instruction tuning phase.
Models with frozen LLM include BLIVA. Although CogVLM does not update its own LLM parameters, it incorporates an additional module into the LLM and updates this module's parameters.
Models that update LLM parameters include LLaVA, Qwen-VL, and MiniGPT-v2.
\subsection{Embodied Agents}
LLM-based agents, especially embodied agents, have demonstrated rapid and significant emergence \citep{DBLP:journals/corr/abs-2309-07864}.
The embodied components enable them to interact with environments by perceiving the surroundings and performing actions.
We focus on embodied agents that perform actions independently rather than relying on external tools.
This survey analyzes embodied agents by the LLM component, tasks, environments, and design strategies, as summarized in Table \ref{tab:Embodied Agents}.
\begin{table*}
	\scriptsize
	\resizebox{\textwidth}{!}{%
		\begin{tabular}{llllrrrr}
			\toprule
			\textbf{Name} & \textbf{Time} & \textbf{LLM} & \textbf{Task} & \textbf{Environment} & \textbf{LM Output} & \textbf{Code Interface} & \textbf{LM Prompting/Tuning} \\
			\midrule
			LM as Planner \citep{DBLP:conf/icml/HuangAPM22} & 2022.01 & GPT3, CodeX & VirtualHome tasks & \ding{192} & \ding{192} & \ding{192} & \ding{192} \\
			SayCan \citep{DBLP:conf/corl/IchterBCFHHHIIJ22} & 2022.04 & PaLM & Real-world robotic tasks & \ding{193} & \ding{193} & \ding{192} & \ding{192} \\
			Code as Policy \citep{DBLP:conf/icra/LiangHXXHIFZ23} & 2022.09 & Codex & Real-world robotic tasks & \ding{193} & \ding{193} & \ding{193} & \ding{192} \\
			Inner Monologue \citep{DBLP:conf/corl/HuangXXCLFZTMCS22} & 2022.07 & InstructGPT & Ravens tasks and real-world robotic tasks & \ding{192} + \ding {193} & \ding{192} & \ding{192} & \ding{192} \\
			ProgPrompt \citep{DBLP:conf/icra/SinghBMGXTFTG23} & 2022.09 & GPT-3 & VituralHome tasks and real-world robotic tasks & \ding{192} + \ding {193} & \ding{193} & \ding{193} & \ding{192} \\
			LLM-Planner \citep{DBLP:journals/corr/abs-2212-04088} & 2022.12 & GPT-3 & Alfred tasks & \ding{192} & \ding{192} & \ding{192} & \ding{192} \\
			Matcha \citep{DBLP:journals/corr/abs-2303-08268} & 2023.03 & GPT-3 & CoppeliaSim-simulated NICOL robot tasks & \ding{192} & \ding{192} & \ding{192} & \ding{192} \\
			PaLM-E \citep{DBLP:conf/icml/DriessXSLCIWTVY23} & 2023.03 & PaLM & TAMP tasks and real-world robotic tasks & \ding{193} & \ding{192} & \ding{192} & \ding{192} \\
			Grounded Decoding \citep{DBLP:journals/corr/abs-2303-00855} & 2023.03 & InstructGPT, PaLM & Ravens tasks and real-world robotic tasks & \ding{192} + \ding {193} & \ding{192} & \ding{192} & \ding{192} \\
			Voyager \citep{DBLP:journals/corr/abs-2305-16291} & 2023.05 & GPT-4 & Minecraft Tasks & \ding{192} & \ding{193} & \ding{193} & \ding{192} \\
			GITM \citep{DBLP:journals/corr/abs-2305-17144} & 2023.05 & GPT-3.5 & Minecraft Tasks & \ding{192} & \ding{193} & \ding{192} & \ding{192} \\
			RT-2 \citep{DBLP:journals/corr/abs-2307-15818} & 2023.07 & PaLI-X,PaLM-E & Real-world robotic tasks & \ding{193} & \ding{193} & \ding{192} & \ding{193}\\
			ViLA \citep{DBLP:journals/corr/abs-2311-17842} & 2023.11 & GPT-4V & Ravens tasks and real-world robotic tasks & \ding{192} + \ding{193} & \ding{192} & \ding{192} & \ding{192} \\
			AutoRT \citep{ahn2024autort} & 2024.01 & / & Real-world robotic tasks & \ding{193} & \ding{192} & \ding{192} & \ding{192} \\
			\bottomrule
		\end{tabular}%
	}
	\caption{
		Summary of LLM-based embodied agents.
		Categorized by environment, \ding{192} symbolizes simulated enrionments and \ding{193} the physical world. 
		In LM output, \ding{192} means mid-level plans, while \ding{193} indicates low-level robotic actions. 
		Regarding code interface, \ding{192} is code-free (language-only), and \ding{193} involves programming code.
		In LM operations, \ding{192} is for prompting, and \ding{193} for tuning.
		It is observed that most embodied agents employ LM prompting for mid-level plans in natural language, which are then translated into low-level robotic actions.}
	\label{tab:Embodied Agents}
\end{table*}

\paragraph{LLM Components.} GPT-3.5 and GPT-4, known for advanced cognitive abilities, are the favored LLMs for developing embodied agents.
Models such as Inner Monologue \citep{DBLP:conf/corl/HuangXXCLFZTMCS22}, ProgPrompt \citep{DBLP:conf/icra/SinghBMGXTFTG23}, LLM-Planner \citep{DBLP:journals/corr/abs-2212-04088}, Matchat \citep{DBLP:journals/corr/abs-2303-08268}, Grounded Decoding \citep{DBLP:journals/corr/abs-2303-00855}, Voyager \citep{DBLP:journals/corr/abs-2305-16291}, GITM \citep{DBLP:journals/corr/abs-2305-17144}, and ViLA \citep{DBLP:journals/corr/abs-2311-17842} utilize GPT models as the cognitive core for thinking and planning.
Additionally, `Code as Policy' \citep{DBLP:conf/icra/LiangHXXHIFZ23} employs CodeX, a derivative of the GPT series focusing on programming.
Another prominent LLM is PaLM, which has been chosen by agents like SayCan \citep{DBLP:conf/corl/IchterBCFHHHIIJ22}, Grounded Decoding, and RT-2 \citep{DBLP:journals/corr/abs-2307-15818}.

\paragraph{Environments and Tasks.}
There are two types of environments in which embodied agents operate: simulated and real-world environments.
Within simulated settings, agents are trained to execute predefined skills.
For instance, Voyager and GITM are designed for playing Minecraft.
Additionally, some agents, functioning as virtual robots, utilize platforms like VirtualHome \citep{DBLP:conf/cvpr/PuigRBLWF018} for cost-effective training prior to potential deployment in real-world scenarios.
Furthermore, numerious embodied agents are able to execute in the physical world.
These include SayCan, `Code as Policy', Inner Monologue, ProgPrompt, Matcha, PaLM-E, Grounded Decoding, RT-2, ViLA, AutoRT \citep{ahn2024autort}, and so on.

\paragraph{LLM-based Agents.}
To transform LLMs into agents, a typical method is a two-tier framework:
the LLM initially generates natural language sequences as high-level plans, later translated into executable low-level actions.
Agents using this approach include Inner Monologue, LLM-Planner, Matcha, PaLM-E, Grounded Decoding, ViLA, and AutoRT.
Alternatively, some models have implemented an end-to-end action generation scheme.
Such agents include SayCan, `Code as Policy', ProgPrompt, Voyager, GITM, and RT-2.
Particularly, RT-2 treats action as a modality and has introduced the concept of a visual-language-action model.
In addition, while many agents leverage the prompt approach in utilizing LLMs, RT-2 fine-tunes the LLMs for improved performance.
Moreover, as next-step action generation is inherently logical, some research conveys a preference for using code instead of natural language to facilitate this process.
Agents that have implemented programming code for action generation include ProgPrompt, `Code as Policy', and Voyager.

\section{\label{sec:benchmark}Benchmark Evaluation}
In order to explore the question `Can LMs transform the landscape of VIA?' and to develop a roadmap for VIALM, we conducted a benchmark evaluation.
This assessment involves measuring and comparing the zero-shot performance of LMs, aiming to help readers understand the models' strengths and weaknesses in the following four key aspects.
(RQ1) \textit{Fundamental Skills}: Do LMs possess the  fundamental multimodal capabilities necessary for VIA?
(RQ2) \textit{Practical Skills}: Can LMs achieve effective performance in practice with complex physical environments?
(RQ3) \textit{Model Selection}: How do different designs of LMs affect performance in VIA?

\paragraph{Models and Setup.}
\citet{DBLP:journals/corr/abs-2310-02071} demonstrated that end-to-end VLMs outperform separate visual and language pipelines in embodied tasks.
We selected six SOTA end-to-end VLMs for our evaluation: GPT-4, CogVLM, MiniGPT, Qwen-VL, LLaVA and BLIVA.
The specific versions were \textit{GPT-4}, \textit{CogVLM-17B-Chat-V1.1}, \textit{MiniGPT-v2 with LLaMA-2-Chat-7B}, \textit{Qwen-VL-Chat with Qwen-7B}, \textit{LLaVA-v1.5-7B}, and \textit{Bliva-vicuna-7B}.
All models operated with float16 precision, with the exception of GPT-4.
We explicitly constructed the prompts, requesting LMs to act as a VI assistant and provide grounded guidance to follow with tactile clues.

\subsection{\label{ssec:eval-dataset}Evaluation Dataset}
Given that VIALM processes environmental information and user requests to ultimately generate step-by-step guidance, the benchmark is structured to follow the VQA framework. 
The input comprises an \textit{image} depicting the scene in front of the VI user, combined with a \textit{question} reflecting their intention.
The output is an \textit{answer} that delivers the desired guidance.
Due to the scarcity of datasets specifically tailored for the VIALM task and the widespread use of popular VQA datasets \citep{DBLP:conf/cvpr/MarinoRFM19, DBLP:conf/cvpr/Gurari0SGLGLB18} in the fine-tuning and instruction tuning of VLMs \citep{DBLP:journals/corr/abs-2305-06500}, we have carefully developed an evaluation dataset. This dataset is designed to (1) accurately reflect, measure, and compare performance on the VIALM task, and (2) remain unexposed to these VLMs during their training and tuning phases.

\paragraph{Dataset Annotation.}

\begin{figure}
	\centering
	\includegraphics[angle=0,width=0.49\textwidth]{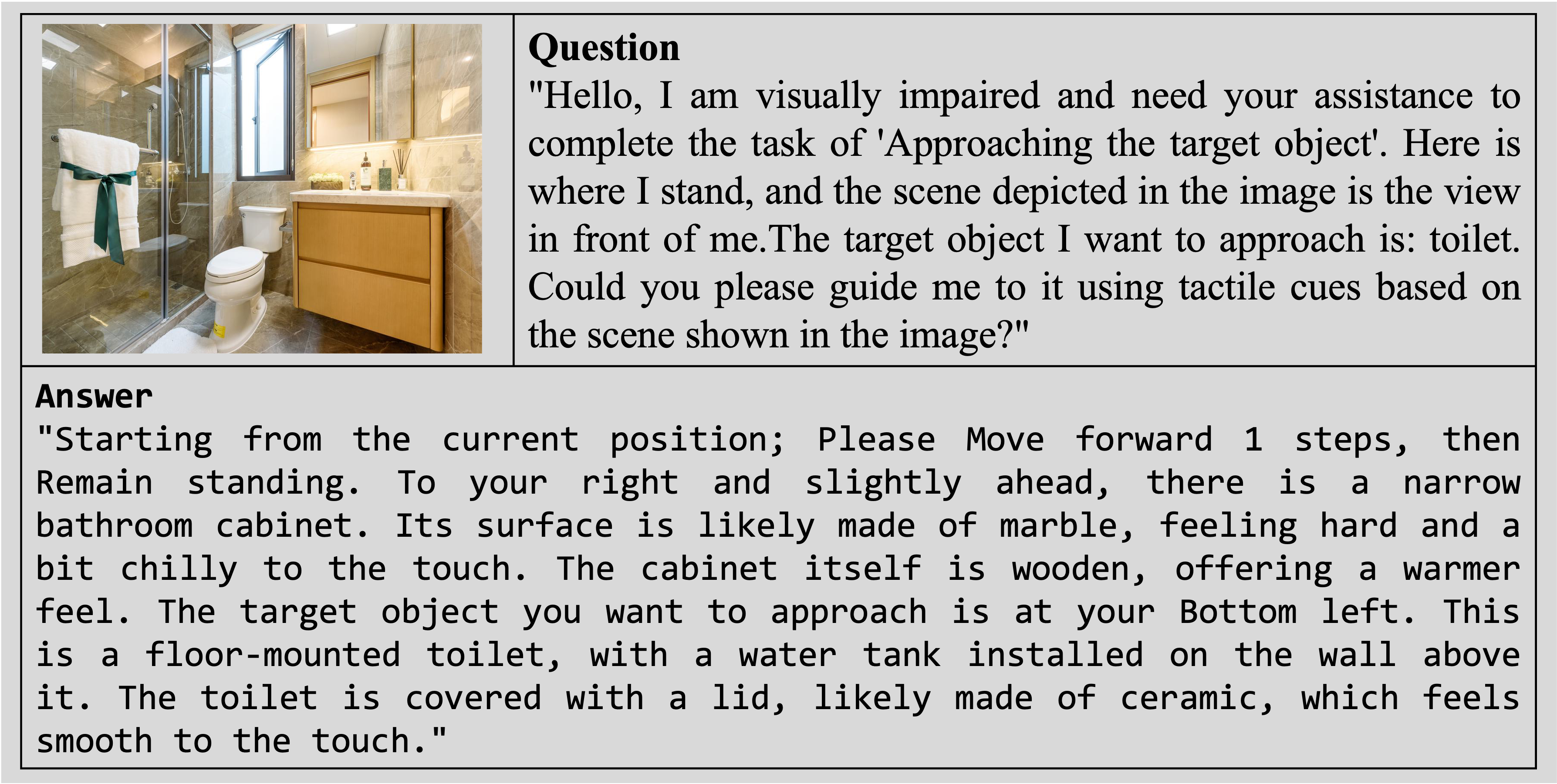}
	\caption{Evaluation Data Sample.The test dataset is in the format of a VQA dataset. The question indicates that the VI user is searching for a toilet. The output answer offers detailed guidance related to this environment to assist the user in achieving this objective.}
	\label{Benchmark Example}
\end{figure}
We collected images from the Internet of the two most common daily life environments: homes and supermarkets.
The dataset was subsequently manually curated by employing well-educated and English-fluent annotators, following the practice in \citet{DBLP:conf/emnlp/LiCZNW23}.
Specifically, the ground truth construction is supervised by VIA experts at The Hong Kong Society for the Blind (HKSB).

\paragraph{Dataset Characteristics.}
Each image in the dataset is paired with a corresponding question and an answer.
Questions are crafted to encourage the models to offer guidance (answers) tailored for VI users.
In formulating answers, tactile sensation is considered, as previous work indicates that tactile perception is more pronounced among the VI group, compensating for vision loss during task completion \citep{ottink2022cognitive}.
%Additional research \citep{israr2012tactile} has also leveraged tactile sensation to develop applications specifically for VI users.
%Consequently, the answers incorporates tactile cues to assist in navigating towards specific targets.
Figure \ref{Benchmark Example} shows an example from the dataset.

\paragraph{Dataset Statistics.}
The evaluation dataset consists of 200 samples in total, divided into two scenarios with 100 images each, similar in size to \citet{DBLP:journals/corr/abs-2310-02071}.
It features a broad \textit{coverage}, with 96 different target objects, such as `meat' and `basin' in the supermarket scenario, and `sofa' and `fridge' in the home scenario.
Furthermore, in comparison to benchmark datasets like OK-VQA (average answer length: 8.1 tokens) \citep{DBLP:conf/cvpr/MarinoRFM19} and CLEVR (average answer length: 18.4 tokens) \citep{DBLP:conf/cvpr/JohnsonHMFZG17}, our dataset exhibits longer questions (average question length: 51 tokens) and significantly longer answers (average answer length: 99.8 tokens).
It indicates a higher level of \textit{difficulty} for models.

\subsection{Evaluation Metrics}
We utilize a hierarchical set of metrics across various dimensions for evaluation. These metrics fall into two main categories: Automatic Evaluation and Human Evaluation.

\paragraph{Automatic Evaluation Metrics.}
We analyze the generated responses \citep{DBLP:conf/acl/NimahFMP23} using both surface-level and semantic-level metrics.
Surface-level metrics evaluate the overlap of tokens or strings between generated and reference texts.
For this, we adopt the ROUGE metric \citep{lin-2004-rouge}, which is the standard for document summarization.
Semantic-level metrics assess the semantic coherence and relevance of the model's outputs in comparison to the ground truths.
For this, we use BERTScore \citep{DBLP:conf/iclr/ZhangKWWA20}, a widely recognized metric that evaluates semantic similarity.

\paragraph{Human Evaluation Metrics.}
The human evaluation measures models' performance in three main aspects:
\textbf{Correctness} assesses whether the guidance can lead the user to complete tasks grounded in environments; 
\textbf{Actionability} measures how easily users can follow the provided guidance, emphasizing the need for guidance to be detailed and clear (\textit{fine-grained}), while also highlighting the importance of \textit{tactile} support for VIA; 
\textbf{Fluency} examines the generation language quality considering the long output in our benchmark ($\S$\ref{ssec:eval-dataset}).
\section{Experiment Results}
\begin{figure}
	\centering
	\includegraphics[angle=0,width=0.5\textwidth]{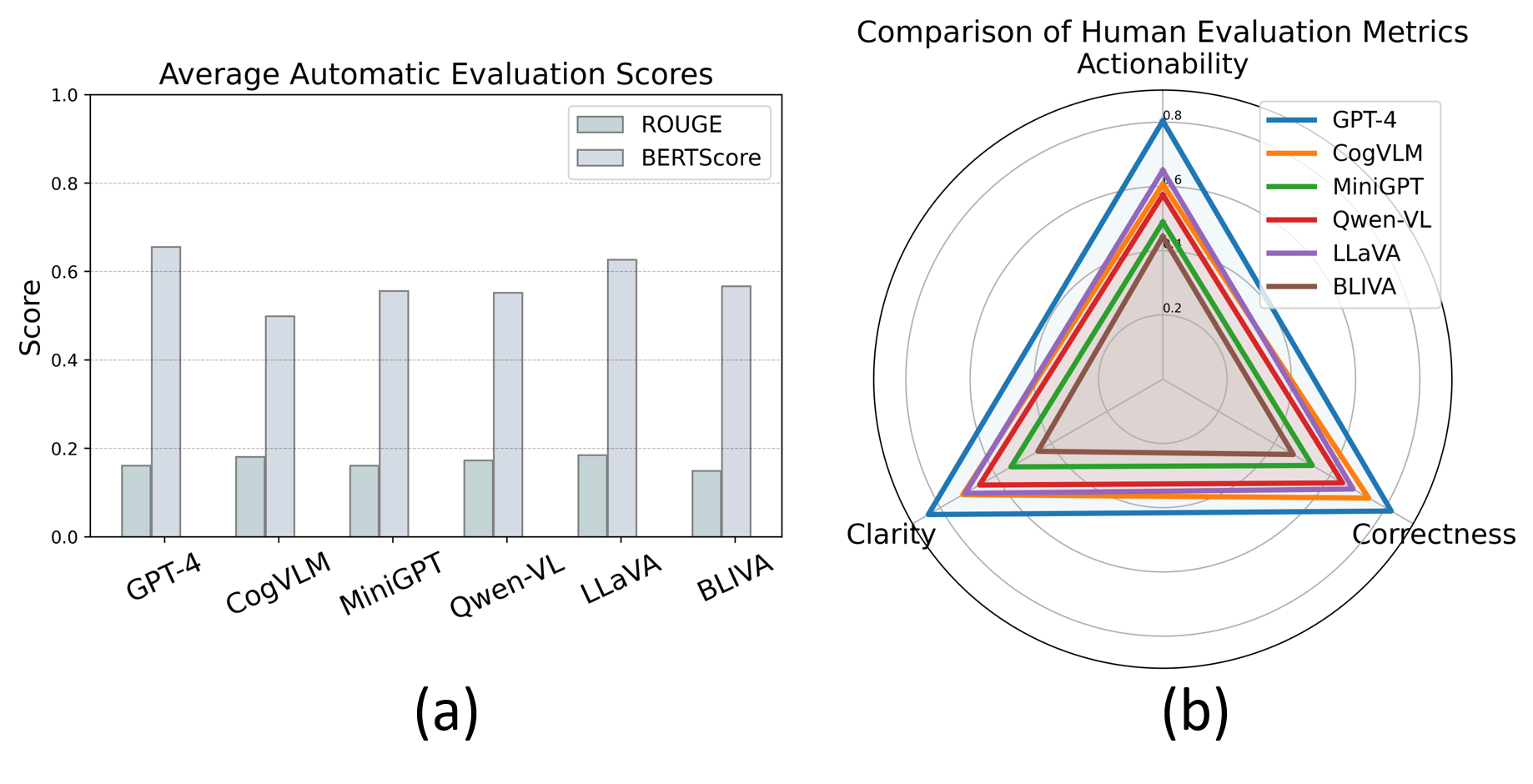}
	\caption{Evaluation Results. (a) The overall average automatic evaluation results for these six models: GPT-4, CogVLM, MiniGPT, Qwen-VL, LLaVA, and BLIVA, based on ROUGE and BERTScore metrics. (b) The human evaluation results of the six models focusing on the aspects of Correctness, Actionability, and Clarity.
	}
	\label{Eval Chat}
\end{figure}
\subsection{Automatic Evaluation Results}
\begin{table}[]
	\footnotesize
	\begin{tabular}{lrrrrr}
		\multicolumn{6}{l}{Supermarket} \\
		\toprule
		\multicolumn{1}{l|}{\textbf{Model}} & \textbf{RG\_1} & \textbf{RG\_2} & \textbf{RG\_L} & \textbf{BERT\_S} & \textbf{Len} \\ \midrule
		\multicolumn{1}{l|}{GPT-4} & 0.246 & 0.047 & 0.138 & \colorbox{lightgray}{\textbf{0.614}} & 229.39 \\
		\multicolumn{1}{l|}{CogVLM} & \colorbox{lightgray}{\textbf{0.319}} & \colorbox{lightgray}{\textbf{0.063}} & \colorbox{lightgray}{\textbf{0.187}} & 0.477 & 65.22 \\
		\multicolumn{1}{l|}{MiniGPT4} & 0.269 & 0.046 & 0.15 & 0.514 & 115.71 \\
		\multicolumn{1}{l|}{Qwen-VL} & 0.254 & 0.05 & 0.156 & 0.505 & 151.31 \\
		\multicolumn{1}{l|}{LLaVA} & \colorbox{lightgray}{0.283} & \colorbox{lightgray}{0.053} & \colorbox{lightgray}{0.168} & \colorbox{lightgray}{0.592} & 139.74\\
		\multicolumn{1}{l|}{BLIVA} & 0.208 & 0.03 & 0.123 & 0.494 & 192.07\\ \hline
		&  &  &  &  & \\
		\multicolumn{6}{l}{Home} \\ \toprule
		\multicolumn{1}{l|}{\textbf{Model}} &\textbf{ RG\_1} & \textbf{RG\_2} & \textbf{RG\_L} & \textbf{BERT\_S} & \textbf{Len} \\ \midrule
		\multicolumn{1}{l|}{GPT-4} & 0.356 & 0.082 & 0.183 & \colorbox{lightgray}{\textbf{0.698}} & 195.07\\
		\multicolumn{1}{l|}{CogVLM} & 0.324 & 0.085 & 0.174 & 0.521 & 52.7\\
		\multicolumn{1}{l|}{MiniGPT4} & 0.331 & 0.081 & 0.182 & 0.597 & 97.85\\
		\multicolumn{1}{l|}{Qwen-VL} & \colorbox{lightgray}{0.357} & \colorbox{lightgray}{0.087} & \colorbox{lightgray}{0.19} & 0.598 & 88.13\\
		\multicolumn{1}{l|}{LLaVA} & \colorbox{lightgray}{\textbf{0.374}} & \colorbox{lightgray}{\textbf{0.101}} & \colorbox{lightgray}{\textbf{0.201}} & \colorbox{lightgray}{0.662} & 103.78\\
		\multicolumn{1}{l|}{BLIVA} & 0.317 & 0.073 & 0.175 & 0.639 & 195.79\\ \bottomrule
	\end{tabular}
	\caption{Automatic Evaluation Results.
		Specific versions are GPT-4, CogVLM-17B-Chat-V1.1, MiniGPT-v2-7B, Qwen-VL-Chat-7B, LLaVA-v1.5-7B, and Bliva-vicuna-7B.
		Except for GPT-4, the others operated with float16 precision.
		The highest and runner-up metric values are distinctly marked in \colorbox{lightgray}{gray}.
		`RG' represents ROUGE, `BERT\_S' stands for BERTScore, and `Len' signifies output length.}
	\label{res:auto_eval_intergral}
\end{table}
The overall automatic experimental results are illustrated in Figure \ref{Eval Chat} (a).
We observe GPT-4 and LLaVA exhibit relatively high BERTSCore values (above 0.6 within a range of 0 to 1).
For ROUGE, all models demonstrate consistently low values (below 0.2 within a range of 0 to 1).
Given that BERTScore quantifies semantic similarity while ROUGE measures strict matching to the references, this observation illustrates that LMs can generally address the task, though without strict matching, indicating LMs' fundamental VIA skills (RQ1).

In a detailed analysis, Table \ref{res:auto_eval_intergral} presents automatic evaluation results for the two distinct scenarios. 
In both situations, GPT-4 and LLaVA possess the highest BERTScores, demonstrating superior semantic capabilities compared to other models.
Additionally, regarding ROUGE, we can see: (1) all models perform better in the home scenario setting; (2) LLaVA displays relatively high ROUGE values and CogVLM excels in a more challenging environment; (3) GPT-4  performs worse than smaller-scaled models.
The first observation suggests that the increased visual complexity of the supermarket environment (i.e., a greater variety of objects and more colors in the image) makes the VIALM task more challenging, highlighting the need for models to generalize with in more challenging environments (RQ2).
Combining the second and third observations and analyzing the generated guidance, we identify GPT-4's significant drawback: its output is excessively redundant and lengthy, resulting in low ROUGE scores.
In contrast, CogVLM generates the shortest outputs, while LLaVA exhibits the best performance with a moderate guidance length and high ROUGE scores.
\begin{table}[]
	\resizebox{\columnwidth}{!}{%
		\begin{tabular}{l|rr}
			\toprule
			\textbf{Model} & \textbf{Not Grounded Rate} & \textbf{Not Fine-grained Rate} \\ \midrule
			GPT-4 & \textbf{0.257} & \textbf{0.321} \\
			CogVLM & 0.45 & 0.721 \\
			MiniGPT4 & 0.736 & 0.8 \\
			Qwen-VL & 0.607 & 0.743 \\
			LLaVA & 0.486 & 0.543 \\
			BLIVA & 0.85 & 0.886 \\ \bottomrule
		\end{tabular}%
	}
	\caption{Human Evaluation Fail Rates.
		`Not Grounded Rate' is calculated by the sample proportion with Correctness $\leq$ 3 (Score range: 1--5).
		`Not Fine-grained Rate' is derived from Actionability $\leq$ 3.}
	\label{rate}
\end{table}

\subsection{\label{ssec:human_eval_res}Human Evaluation Results}
Figure \ref{Eval Chat} (b) presents the human evaluation results.
To mitigate bias, two independent participants conducted evaluations.
In each scenario, 50 random samples were selected.
Pearson and Spearman correlation coefficients between the two annotators for each dimension range from 0.7 to 0.9, indicating a strong consensus on the models' performance. 
First, it is once again observed that all models provide more accurate answers in home settings than in supermarket settings.
Second, as shown in Figure \ref{Eval Chat} (b), GPT-4 excels in performance across all three dimensions, with CogVLM and LLaVA closely following in second place.
Subsequently, we analyze the results for the three specific dimensions: 

$\bullet$ In Correctness, GPT-4 performs the best, with CogVLM slightly surpassing LLaVA.
It indicates that CogVLM understands the visual environment better than LLaVA, and therefore leads to a superior \textit{environment grounding} ability.

$\bullet$ In Actionability, GPT-4 excels, with LLaVA narrowly outperforming CogVLM.
It suggests LLaVA's LLM component, LLaMA-2, allows for more \textit{fine-grained} and user-friendly guidance.
As we explicitly prompt for tactile clues ($\S$\ref{sec:benchmark}), we observe that only GPT-4 can provide some general tactile information, while the other models cannot provide.

$\bullet$ In Clarity, GPT-4 obtains the highest score.
Nevertheless, its responses tend to be redundant and long-winded, necessitating extra effort to achieve functionality.
In contrast, other models provide responses that are not excessively long but tend to be overly simplistic, lacking sufficient information.

This analysis reveals \textit{two limitations shared by all models}, summarized as the answer to RQ2:  `\textbf{not grounded}' (indicated by low correctness) and `\textbf{not fine-grained}' (indicated by low actionability).
Table \ref{rate} presents the failure rates.
It is notable that, despite GPT-4's superior performance, 25.7\% of its responses are not grounded and 32.1\% not fine-grained.
Additionally, only GPT-4 with a very large scale is capable of providing some general tactile descriptions, while the other five smaller models largely lack this ability.

\paragraph{Case Study.}
\begin{figure}
	\centering
	\includegraphics[angle=0,width=0.49\textwidth]{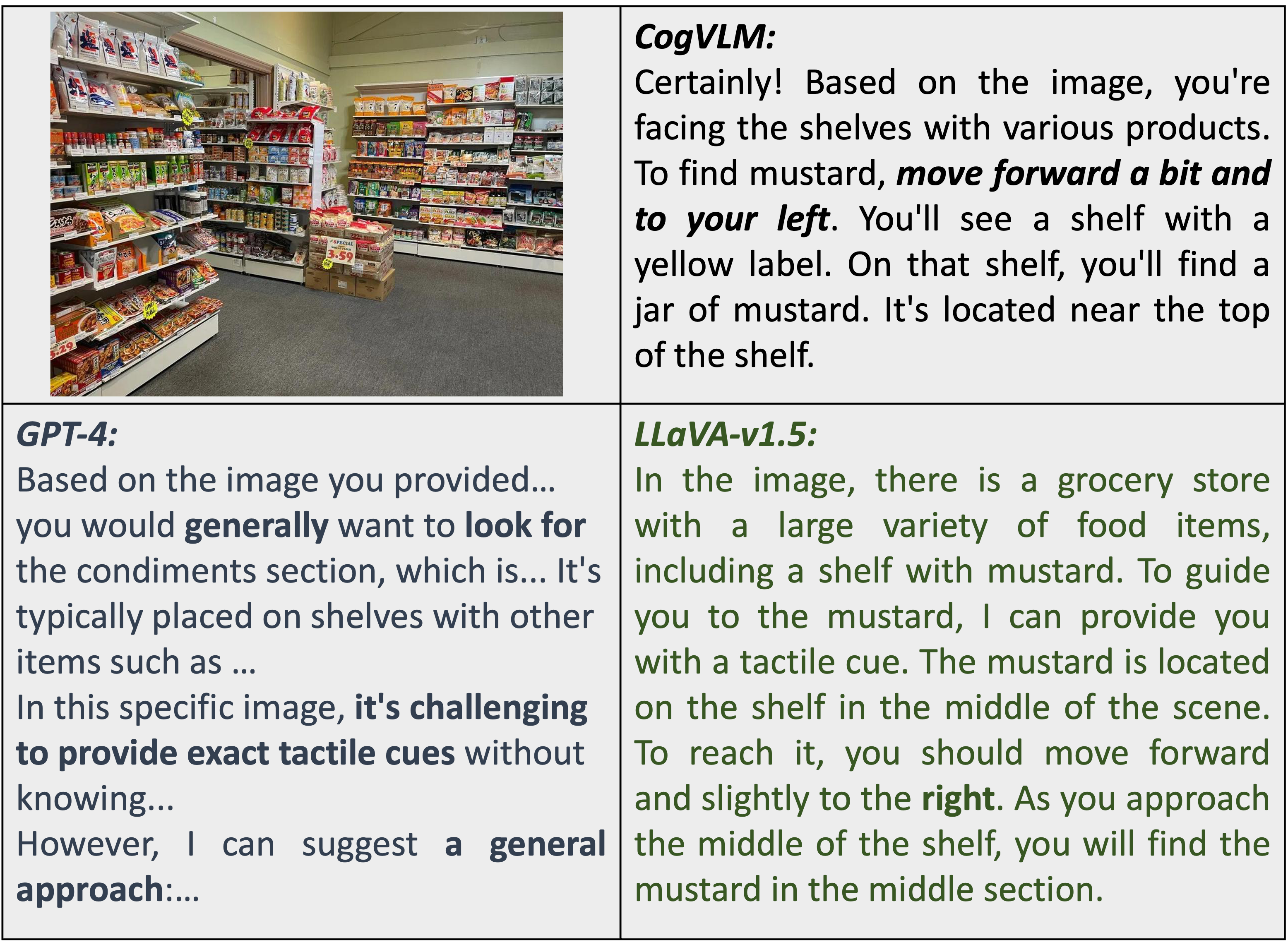}
	\caption{An example of predictions from the top three models. GPT-4 is prone to generating redundant output describing general situations. CogVLM exhibits advantages in image grounding, while LLaVA emphasizes providing step-by-step guidance for VI users. }
	\label{case}
\end{figure}
To better understand the above results, we present an example case in Figure \ref{case} where the question seeks guidance to obtain mustard at a supermarket.
The predictions of GPT-4, CogVLM, and LLaVA yield the following observations:
(1) GPT-4 generates long output with general guidance but lacks image grounding.
(2) CogVLM exhibits better environment grounding, correctly specifying the object's position, yet it lacks detailed step-by-step guidance.
(3) LLaVA's output emphasizes guiding the user to the specific object but appears to have limited understanding of the environment.
(4) GPT-4 generates general tactile guidance, whereas CogVLM and LLaVA entirely ignore tactile descriptions.
\subsection{Discussions}

\begin{table}[ht]
	\centering
	\scriptsize
	\resizebox{\columnwidth}{!}{%
		\begin{tabular}{lrrrr}
			\toprule
			& \multicolumn{2}{c}{\textbf{Grounded}} & \multicolumn{2}{c}{\textbf{Fine-grained}} \\
			\cmidrule(r){2-3} \cmidrule(r){4-5}
			\textbf{Model} & \textbf{RG\_L} & \textbf{BERT\_S} & \textbf{RG\_L} & \textbf{BERT\_S} \\
			\midrule
			Supermarket & & & & \\
			GPT-4 & 0.072 & 0.245 & 0.115 & \textbf{0.588} \\
			CogVLM & \textbf{0.144} & 0.14 & \textbf{0.172} & 0.495 \\
			MiniGPT & 0.102 & 0.202 & 0.136 & 0.485 \\
			Qwen-VL & 0.09 & 0.169 & 0.138 & 0.495 \\
			LLaVA & 0.097 & \textbf{0.26} & 0.145 & 0.554 \\
			BLIVA & 0.062 & 0.213 & 0.107 & 0.463 \\
			\midrule
			Home & & & & \\
			GPT-4 & 0.104 & 0.405 & 0.166 & \textbf{0.573} \\
			CogVLM & \textbf{0.158} & 0.284 & 0.189 & 0.443 \\
			MiniGPT & 0.139 & \textbf{0.415} & 0.177 & 0.455 \\
			Qwen-VL & 0.132 & 0.311 & 0.193 & 0.506 \\
			LLaVA & 0.123 & 0.404 & \textbf{0.202} & 0.544 \\
			BLIVA & 0.091 & 0.375 & 0.159 & 0.514 \\
			\bottomrule
		\end{tabular}%
	}
	\caption{Automatic Evaluation Results with separated guidance parts extracted from ground truths. Metric scores are calculated by comparing model predictions with each corresponding part.}
	\label{tab:auto_eval_home_sep}
\end{table}
%two scenes
This section explores model selection (RQ3) strategies to improve VIALM, particularly in addressing the two mentioned limitations ($\S$\ref{ssec:human_eval_res}).
We conducted additional experiments, dividing the ground truth text into two parts: grounded and fine-grained.
We then compared predictions with each part and calculated automatic metric scores, as displayed in Table \ref{res:auto_eval_home_sep}.

It is observed that CogVLM performs well in environment grounding and LLaVA excels in generating fine-grained outputs. 
CogVLM's success in environmental grounding might stem from two factors: (1) the incorporation of the more recent Eva-Clip ViT model, enhancing visual perception, and (2) the inclusion of a Visual Expert module in CogVLM's LLM architecture, augmenting visual-language alignment.
LLAVA's effectiveness in producing fine-grained guidance might come from its advanced LLM component, LLaMA-2.

Enhancing environment grounding demands advanced visual capabilities, and generating fine-grained guidance necessitates better language abilities.
Existing models are limited in synergizing both vision and language skills.
It requires future efforts to develop LMs with environment-grounded multimodal abilities for providing fine-grained guidance in VIA.
\section{Conclusion}
% Add one sentence describing the motivation
In this study, we defined the task of VIALM and carried out a comprehensive survey of LMs for VIA, encompassing LLMS, VLMs, and embodied agents.
Furthermore, we constructed a benchmark dataset with comprehensive coverage and high difficulty.
Following this, we conducted a benchmark evaluation to measure the zero-shot performance of LMs, aiming to examine the potential and limitations of current SOTA models in the VIALM task.
The experimental results revealed two primary limitations in the models: their ability to generate \textit{grounded} and \textit{fine-grained} guidance.
\bibliographystyle{named}
\bibliography{ijcai24}

\end{document}